\documentclass{article}
\usepackage{spconf,graphicx}
\usepackage{amsmath,amsfonts,bm,multicol,multirow}
\usepackage{hyperref}
\usepackage{xcolor}
\usepackage{cleveref}
\usepackage{tabularx}
\usepackage{booktabs}
\usepackage{mathtools}
\usepackage[toc,page]{appendix}
\usepackage{subcaption}

\usepackage{microtype}

\usepackage{latexsym}

\newcommand{\R}{\mathbb{R}}

\usepackage[scientific-notation=true]{siunitx}
\usepackage{amsmath}

\DeclareMathOperator*{\Exp}{\mathbb{E}}

\newcommand{\Inp}{\mathcal{X}}
\newcommand{\Feat}{\mathcal{Z}}
\newcommand{\Context}{\mathcal{C}}

\newcommand{\z}{\mathbf{z}}
\newcommand{\tz}{\mathbf{\tilde{z}}}
\newcommand{\cc}{\mathbf{c}}

\newcommand{\ze}{\mathbf{z}}
\newcommand{\zq}{\mathbf{\hat{z}}}

\newcommand{\e}{\mathbf{e}}

\newcommand{\Enot}[1]{\num[exponent-product = \times]{#1}}

\makeatletter
\renewcommand*{\@fnsymbol}[1]{\ensuremath{\ifcase#1\or * \or \dagger\or \ddagger\or
    \mathsection\or \mathparagraph\or \|\or **\or \dagger\dagger
    \or \ddagger\ddagger \else\@ctrerr\fi}}
\makeatother

\usepackage{soul} 

\newif\ifdraftmode
\draftmodetrue

\ifdraftmode

\newcommand{\comment}[1]{\textcolor{red}{TODO: \emph{#1}}}

\else

\newcommand{\comment}[1]{}

\fi

\usepackage[small,compact]{titlesec} 
\usepackage{times}
\title{Effectiveness of self-supervised pre-training for speech recognition} 

\name{Alexei Baevski, Michael Auli,  Abdelrahman Mohamed}
\address{Facebook AI Research\\
\texttt{\{abaevski, michaelauli, abdo\}@fb.com} \\
}

\begin{document}

\maketitle
\begin{abstract}


We compare self-supervised representation learning algorithms which either explicitly quantize the audio data or learn representations without quantization.
We find the former to be more accurate since it builds a good vocabulary of the data through vq-wav2vec \cite{baevski2019vqwav2vec} to enable learning of effective representations in subsequent BERT training. 
Different to previous work, we directly fine-tune the pre-trained BERT models on transcribed speech using a Connectionist Temporal Classification (CTC) loss instead of feeding the representations into a task-specific model. 
We also propose a BERT-style model learning directly from the continuous audio data and compare pre-training on raw audio to spectral features.
Fine-tuning a BERT model on 10 hour of labeled Librispeech data with a vq-wav2vec vocabulary is almost as good as the best known reported system trained on 100 hours of labeled data on test-clean, while achieving a 25\% WER reduction on test-other. When using only 10 minutes of labeled data, WER is 25.2 on test-other and 16.3 on test-clean. 
This demonstrates that self-supervision can enable speech recognition systems trained on a near-zero amount of transcribed data.
\end{abstract}

\section{Introduction}
\label{intro}

Representation learning has been an active research area for more than 30 years \cite{hinton_86}, with the goal of learning high level representations which separates different explanatory factors of the phenomena represented by the input data \cite{DL_nature, yoshua_2012}. 
Building Automatic Speech Recognition (ASR) systems, typically requires a large volume of training data to represent different factors contributing to the creation of speech signals, e.g. background noise, recording channel, speaker identity, accent, emotional state, topic under discussion, and the language used in communication. The practical need for building ASR systems for new conditions with limited resources spurred a lot of work focused on unsupervised speech recognition and representation learning~\cite{pg_08, aren_10, glass_12, JHU_2012, JSALT_2017, s2v, cpc, schneider2019wav2vec}, in addition to semi- and weakly-supervised learning techniques to reduce the supervised data needed in real-world scenarios \cite{Vesely_13, Sheng_2017, hari_2019, tara_sslearning, Alishahi_2017, ttic_may2017}.

Recently impressive results have been reported for representation learning, that generalizes to different downstream tasks, through self-supervised learning for NLP and speech \cite{bert, baevski2019cloze, cpc, schneider2019wav2vec, baevski2019vqwav2vec}. Self-supervised representation learning tasks include predicting masked parts of the input, reconstructing inputs through low bit-rate channels, or contrasting similar data points against different ones. 

In this work we compare different approaches of self-supervised pre-training for speech data. 
We consider learning discrete units to represent the audio data through either self-supervision~\cite{baevski2019vqwav2vec} or through clustering spectral features, followed by pre-training over these units using a bi-directional transformer (BERT) model~\cite{bert}.
This is compared to directly learning representations without explicit quantization over the raw audio as well as spectral features.
Previous work fed the representations of the pre-trained model into a task-specific architecture for speech recognition instead of the raw waveform~\cite{schneider2019wav2vec,baevski2019vqwav2vec}.
Instead we directly fine-tune the pre-trained BERT model on transcribed speech data using a CTC loss.
Our experiments demonstrate that discrete  unit discovery, followed by BERT training achieves better results than representations learned without explicit quantization. Disentangling acoustic unit discovery from learning the sequential relationship between them, enables better representations of the data which in turn improves down-stream model accuracy.

We pre-train our models on the unlabeled 960h Librispeech~\cite{librispeech} data and follow the Libri-light~\cite{librilight} limited resource supervised training sets of 10 hours, 1 hour, and 10 mins. Our best model fine-tuned only on 1 hour of labeled data can outperform the best known result from the literature relying on 100h of labeled data~\cite{L_scher_2019} on the standard Librispeech test-other subset. Using only 10 minutes of labeled data the approach achieves 16.3/25.2 WER on test-clean/other.

\section{Preliminaries}
\subsection{BERT}
Using self-supervision, BERT \cite{bert}, a deep bidirectional transformer model, builds its internal language representation that generalizes to other downstream NLP tasks. Self-attention over the whole input word sequence enables BERT to jointly condition on both the left and right context of data. For training, it uses both a masked language model loss, by randomly removing some input words for the model to predict, and a contrastive loss to distinguish the next sentence in the document from a randomly selected one. 

\subsection{Wav2Vec}
\label{sec:wav2vec}

Wav2Vec~\cite{schneider2019wav2vec} learns representations of audio data by solving a self-supervised context-prediction task with the same loss function as word2vec~\cite{mikolov2013word2vec,cpc}. The model is based on two convolutional neural networks where the \emph{encoder} $f: \Inp \mapsto \Feat$ produces a representation $\z_{i}$ for each time step \emph{i} at a rate of 100~Hz and the \emph{aggregator} $g: \Feat \mapsto \Context$ combines multiple encoder time steps into a new representation $\cc_i$ for each time step \emph{i}.
Given $\cc_i$, the model is trained to distinguish a sample $\z_{i+k}$ that is k steps in the future from distractor samples $\tz$ drawn from a distribution $p_n$, by minimizing the contrastive loss for steps $k=1,\dots,K$:
\begin{align}
\begin{split}
  \mathcal{L}_k = - \sum_{i=1}^{T - k} \Big(
  &\log \sigma(\z_{i+k}^\top h_k(\cc_i)) \\
  &+\lambda \Exp_{\mathclap{\tz \sim p_n}}\ [ \log \sigma(-\tz^\top h_k(\cc_i)) ]\ \Big)\,
  \label{eq:old-objective}
\end{split}
\end{align}
where $T$ is the sequence length, $\sigma(x) = 1/(1+\exp(-x))$, and where $\sigma(\z_{i+k}^\top h_k(\cc_i))$ is the probability of $\z_{i+k}$ being the true sample. A step-specific affine transformation $h_k(\cc_i) = W_k \cc_i + \mathbf{b}_k$ is applied to $\cc_i$~\cite{cpc}.
The loss $\mathcal{L} = \sum_{k=1}^K \mathcal{L}_k$ is optimized by summing~(\ref{eq:old-objective}) over different step sizes.
The learned high level features produced by the context network $\cc_i$ are shown to be better acoustic representations for speech recognition compared to standard spectral features.

\subsection{vq-wav2vec}

vq-wav2vec \cite{baevski2019vqwav2vec} learns vector quantized (VQ) representations of audio data using a future time-step prediction task.
Similar to Wav2Vec, there are a convolutional encoder and decoder networks $f: \Inp \mapsto \Feat$ and $g: \hat{\Feat} \mapsto \Context$ for feature extraction and aggregation. 
However, in between them there is a \emph{quantization} module $q: \Feat \mapsto \hat{\Feat}$ to build discrete representations which are input to the aggregator.

First, 30ms segments of raw speech are mapped to a dense feature representation $\ze$ at a stride of 10ms using the encoder $f$.
Next, the quantizer (q) turns these dense representations into discrete indices which are mapped to a reconstruction $\zq$ of the original representation $\z$.
The $\zq$ is fed into the aggregator $g$ and the model is optimized via the same context prediction task as wav2vec (cf.~\textsection\ref{sec:wav2vec}).
The quantization module replaces the original representation $\ze$ by $\zq = \e_i$ from a fixed size codebook $\e \in \R^{V \times d}$ which contains $V$ representations of size $d$. 

\section{Approach} \label{vqi}
\begin{figure*}
    \centering
    \begin{subfigure}[t]{0.3\textwidth}
        \includegraphics[width=\textwidth,height=6cm]{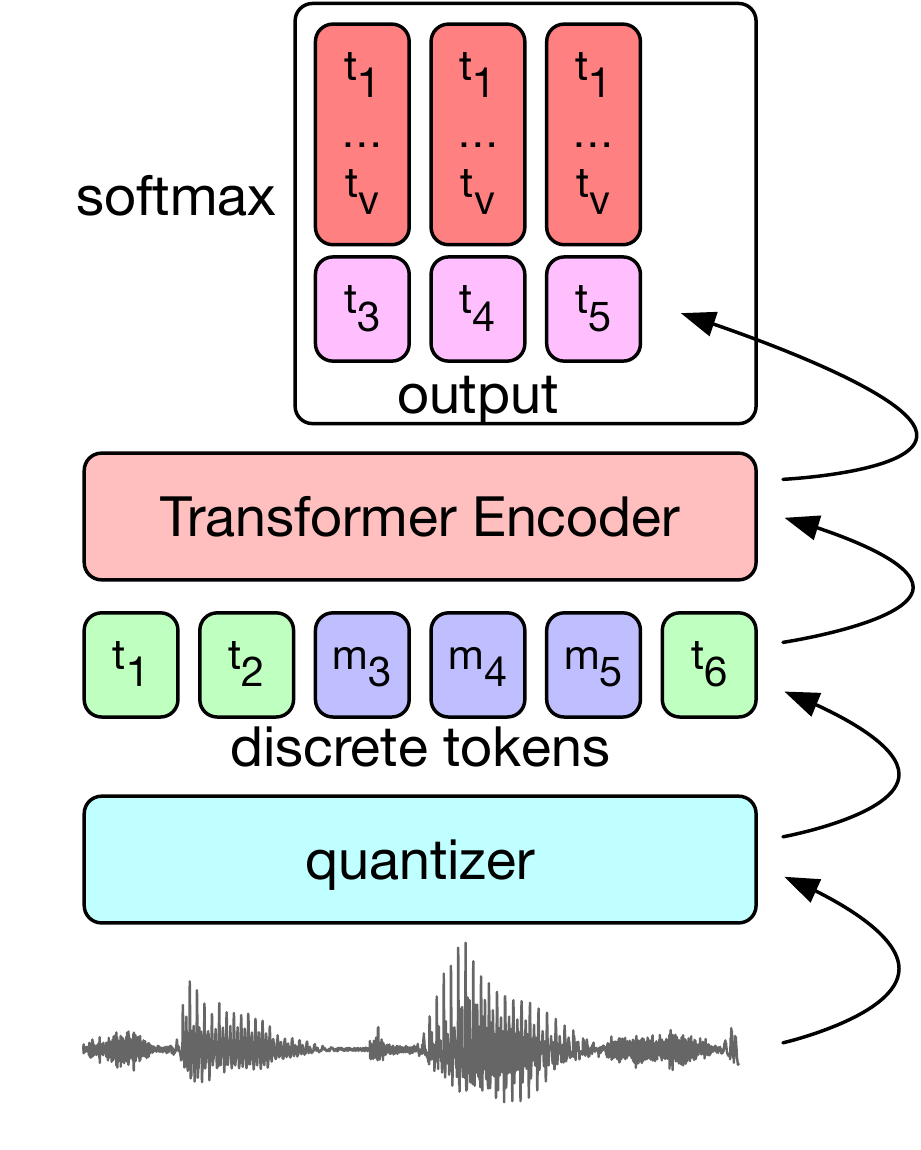}
        \caption{Quantized Inputs
        }
        \label{fig:quant}
    \end{subfigure}
    \hspace{1in} 
    \begin{subfigure}[t]{0.3\textwidth}
        \includegraphics[width=\textwidth,height=6cm]{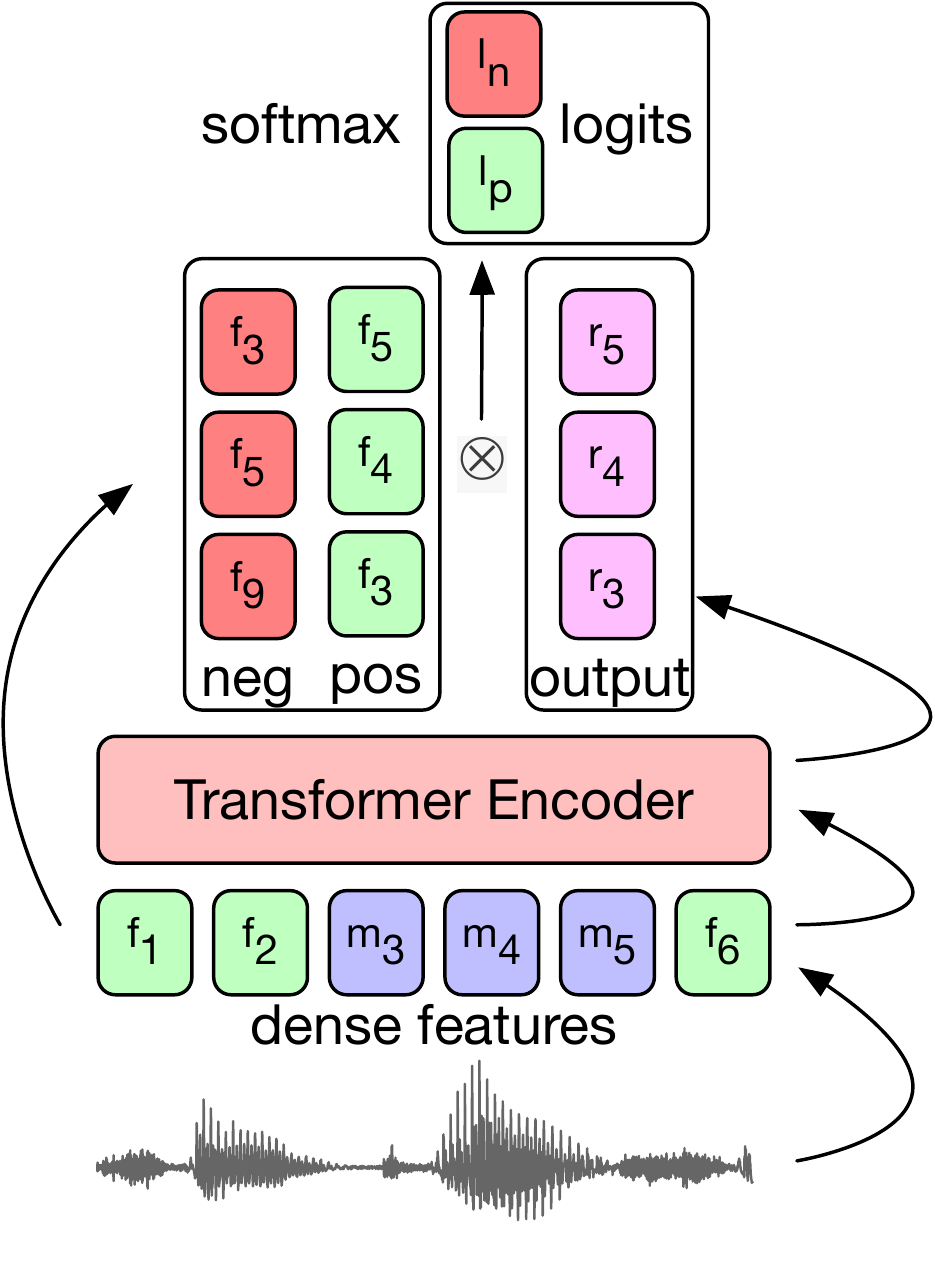}
        \caption{Continuous Inputs
        }
        \label{fig:cont}
    \end{subfigure}
    \caption{
    Illustration of BERT pre-training. $m_i$ refers to masked time-steps chosen at random during each forward pass.
    (\subref{fig:quant}) Inputs are quantized with a vq-wav2vec quantizer or, for MFCC and FBANK variants, by finding the closest centroids and are then used for training a BERT model with a masked language model objective.
    (\subref{fig:cont}) wav2vec, MFCC or FBANK inputs are masked, encoded by a transformer, and then fed into a classifier to predict which features were masked at each time-step.
    }
\end{figure*}
\subsection{Discrete BERT} \label{discrete_bert}

Our work builds on the recently proposed work in \cite{baevski2019vqwav2vec} where audio is quantized using a contrastive loss, then features learned on top by a BERT model \cite{bert}. For the vq-wav2vec quantization, we use the gumbel-softmax variant with the same setup as described in \cite{baevski2019vqwav2vec}. This model quantizes the Librispeech dataset into 13.5k unique codes.

To understand the impact of discrete acoustic representations of vq-wav2vec \cite{baevski2019vqwav2vec}, as alternatives, we explore quantizing the standard mel-frequency cepstral coefficients (MFCC) and log-mel filterbanks coefficients (FBANK), choosing a subset small enough to fit into GPU memory and running k-means with 13.5k centroids (to match the vq-wav2vec setup) to convergence. We then assign the index of the closest centroid to represent each time-step. 

We train a standard BERT model \cite{bert,liu2019roberta} with only the masked language modeling task on each set of inputs in a similar way as described in \cite{baevski2019vqwav2vec}, namely by choosing tokens for masking with probability of 0.05, expanding each chosen token to a span of a length sampled from a normal distribution with mean 10 and standard deviation 10 (spans may overlap) and then computing a cross-entropy loss which attempts to maximize the likelihood of predicting the true token for each one that was masked (Figure \ref{fig:quant}).

Following \cite{conv_trans}, we replace the fixed positional embeddings in the BERT model with a single group convolutional layer that is applied directly on the embeddings before any of the transformer blocks. The convolutional layer has a kernel size of 128 and group size of 16 to reduce the number of added parameters.

\subsection{Continuous BERT}

A masked language modeling task cannot be performed with continuous inputs and outputs, as there are no targets to predict in place of the masked tokens. Instead of reconstructing the input as in \cite{oord2017neural}, we classify the masked positive example among a set of negatives. 
The inputs to the model are dense wav2vec features \cite{schneider2019wav2vec}, MFCC or FBANK features representing 10ms of audio data. Some of these inputs are replaced with a \textit{mask} embedding and are then fed into a transformer encoder. We then compute the dot product between the outputs corresponding to each masked input, the true input that was masked, and a set of negatives sampled from other masked inputs within the same batch. The model is optimized with the InfoNCE loss \cite{cpc} where given one 
positive sample $\z_i$ and $N$ negative samples $\tilde{\z}$ we minimize:
\begin{equation}
    \mathcal{L}_k = \sum_{i=1}^{T} \frac{\exp(\z_i)}{\sum_{j = 1}^{N} \exp(\tilde{\z}^{j})}
\end{equation}
where each sample $\z_i$ is computed as a dot product of the output of the model at timestep $i$ and the true unmasked value of positive example at timestep $i$ or a randomly sampled negative example. To stabilize training, we add the squared sum of logits produced by the dot-product to the loss, and then apply a soft clamp $\hat{s_i}=\lambda\tanh(s_i/\lambda)$ for each logit $s_i$ to prevent the model's tendency to continually increase the magnitude of logits during training~\cite{bachman2019learning}. We use the same kind of convolutional positional layer as described in section \ref{discrete_bert}.

\subsection{Supervised fine-tuning}

The pre-trained models are fine-tuned to perform the ASR task by adding a randomly initialized linear projection on top of the features computed by the transformer models into $V$ classes representing the vocabulary of the task. 
The vocabulary is 29 tokens for character targets plus a word boundary token. The models are optimized by minimizing the CTC loss. We apply SpecAugment \cite{Park_2019} inspired masking to time-steps and channels during training which delays overfitting and significantly improves the final accuracy numbers, especially on the smallest subsets.

We train a single seed for all subsets except the 10 minute one, where we train 5 seeds and choose the best one. For 10 minute subset, some of the seeds fail by entering the overfit regime very early. We train on a single GPU using the Adam optimizer with a tri-state learning scheduler where the learning rate is linearly increased from 1e-7 to 2e-05 in the first stage, held at 2e-5 in the second, and finally linearly decayed to 0 in the third. For 1h and 10min subsets we train for 1250 / 6600 / 12150 updates in each respective stage, for 10h subset we train for 5000 / 16500 / 28500 updates and for 100h we train for 8000 / 52800 / 91200 updates. We use a batch size of 6144 timesteps (61.44 seconds worth of audio) for the 100 hour subset and 3072 timesteps for other subsets.

During fine-tuning, we apply a modified SpecAugment \cite{Park_2019} policy, where we randomly choose a number of starting timesteps to mask, with each timestep having a chance of 3.75\% of being chosen. A span of 20 timesteps starting at each of the chosen position is then replaced with the mask embedding used during unsupervised training (spans may overlap). We also apply channel masking, in which we choose the starting channel index from all channels with a probability of 0.4\% and the length of the channel mask by sampling from a normal distribution with a mean of 64 and standard deviation of 64. The chosen (and possibly overlapping) spans of channels are then zeroed out.

We apply a dropout at every layer of the transformer of 0.1 for 10 minute and 1 hour setup, but we disable it for other subsets as masking described above appears to provide enough regularization.






\section{Experiments}

We implement our models in the fairseq~\cite{ott2019fairseq} toolkit.

\subsection{Data}

All experiments are performed by pre-training on the 960 hours of audio only data of the Librispeech~\cite{librispeech} training set, fine-tuning on the Libri-light \cite{librilight} limited resource supervised training sets of 10 hours (24 speakers), 1 hour (24 speakers), and 10 minutes (4 speakers). The Libri-light training sets are sampled equally from the two clean and noisy portions, a balance of male and female speakers. We also report results of models fine-tuned on 100 hours following the ``train-clean-100'' subset. All models are evaluated on the standard Librispeech dev and test splits.

\subsection{Models}
\subsubsection{Quantized Inputs Training} \label{vq-train}
We first train the vq-wav2vec quantization model following the gumbel-softmax setup described in \cite{baevski2019vqwav2vec}.
After training this model on 960h of Librispeech and quantizing the training dataset, we are left with 13.5k unique codewords combinations. 

For quantizing MFCC and FBANK features extracted using the Kaldi \cite{Povey_ASRU2011} toolkit, we use 8 Volta GPUs with 32GB memory to compute 13.5k K-Means centroids matching the number of unique tokens produced by the vq-wav2vec model. To fit into GPU memory, we subsample 50\% of MFCC features and 25\% of FBANK features from the training set before running the clustering algorithm.

The model we use for the masked language modeling task is a standard BERT model with 12 transformer layers, model dimension 768, inner dimension (FFN) 3072 and 12 attention heads~\cite{bert}.
The learning rate is warmed up over the first 10,000 updates to a peak value of \Enot{1e-5}, and then linearly decayed over a total of 250k updates.
We train on 128 GPUs with a batch size of 3072 tokens per GPU giving a total batch size of 393k tokens~\cite{ott2018scaling} where each token represents 10ms of audio data.

To mask the input sequence, we follow~\cite{baevski2019vqwav2vec} and randomly sample $p=0.05$ of all tokens to be a starting index, without replacement, and mask $M$ consecutive tokens from every sampled index; spans may overlap. $M$ is sampled from a Gaussian distribution with $\mu=10$ and $\sigma=10$, rounded to the nearest integer greater than or equal to zero.

Different from \cite{baevski2019vqwav2vec}, we do not concatenate different utterances to form examples for training, instead each utterance is treated as a single example, as we find that this approach produces better results after fine-tuning.

\subsubsection{Continuous Inputs Training}

For training on dense features, we use a model similar to a standard BERT model with the same parameterization as the one used for quantized input training, but we use the wav2vec, MFCC or FBANK inputs directly. We add 128 relative positional embeddings at every multi-head attention block~\cite{dai2019transformerxl} instead of fixed positional embeddings to make it easier to handle longer examples. We train this model on 8 GPUs with a batch size of 9,600 inputs per GPU, resulting in a total batch size of 76,800. We find that increasing the number of GPUs (which increases the effective batch size) does not lead to better results with this particular setup.

Wav2vec features are 512-dimensional, while MFCC features have 39 dimensions and FBANK features have 80. We introduce a simple linear projection from the feature dimension to BERT dimension (768) for all models.

Similar to~\ref{vq-train}, we mask time-steps by randomly sampling, without replacement, $p=0.05$ of all time-steps to be a starting index, and mask $M$ consecutive time-steps from every sampled index; spans may overlap, where $M$ is sampled from a Gaussian distribution with $\mu=10$ and $\sigma=10$, rounded to the nearest integer greater than or equal to zero.
We sample $10$ negative examples from other masked time-steps from the same example, and an additional $10$ negative examples from masked time-steps occurring anywhere in the batch. We compute a dot product between the original features and the output corresponding to the same time-step after they are processed by the BERT model. We add the squared sum of logits from these computations multiplied by $\lambda=0.04$ to the loss, and then apply a smooth clamp by recomputing each logit $\hat{s_i}=20\tanh(s_i/20)$.

The learning rate is warmed up over the first 10,000 updates to a peak value of \Enot{1e-5}, and then linearly decayed over a total of 250k updates.

\subsection{Methodology}

For quantized inputs, we compute token indices using the gumbel-softmax based vq-wav2vec model. For MFCC and FBANK features we take the index of the closest centroid, as measured by finding the minimum Euclidean distance, to each corresponding feature in the Librispeech dataset. We then train a BERT model as descirbed in~\textsection\ref{vq-train}.

For wav2vec continuous inputs, we use features extracted by the publicly available wav2vec \cite{schneider2019wav2vec} model which contains 6 convolutional blocks in the feature extractor and 11 convolutional blocks in the aggregator module. We use the outputs of the aggregator as features. For MFCC and FBANK, we use those features directly after applying a single linear projection to upsample them to the model dimensionality.

We fine-tune our pre-trained models on either 100 hours of Librispeech train-clean-100 subset, 10 hours, 1 hour, or 10 minutes of labelled data following the Libri-light \cite{librilight} limited training sets. 
We use the standard CTC loss and train for up to 20k updates. We find that the pre-trained models converge after only around 4k updates, while the models trained from scratch tend to converge much later, around 18k updates. We fine-tune all models with a learning rate of $0.0001$ that is linearly warmed up over the first 2k updates and then annealed following a cosine learning rate schedule over the last 18k updates. We set the dropout of the pre-trained BERT models to 0.1 and sweep on dropout of the BERT model outputs before the final projection layer over values between 0.0 and 0.4 in increments of 0.1. For each model, we choose a single best checkpoint that has the best loss on the validation set, which is a combination of dev-clean and dev-other standard Librispeech splits.

\begin{table}[ht!]
\centering 
\setlength\tabcolsep{3.3pt}
\begin{tabular}{ccccccc}
\toprule
\multirow{2}{*}{Model} & Input & \multicolumn{2}{c}{dev} && \multicolumn{2}{c}{test}\\
\cline{3-4}\cline{6-7}
{} & features & clean & other && clean & other \\
\midrule
\multicolumn{7}{c}{\textbf{10 mins of labeled data}}\\
\midrule
\multirow{3}{*}{Discrete BERT} & vq-wav2vec & \textbf{15.7} & \textbf{24.1} && \textbf{16.3} & \textbf{25.2} \\
{} & MFCC & 30.3 & 48.8 && 31.5 & 49.1 \\
{} & FBANK & 35.7 & 53.9 && 35.5 & 55.0 \\
\midrule
\multirow{3}{*}{Continuous  BERT} & wav2vec & 49.1 & 66.0 && 49.5 & 66.3 \\
{} & MFCC & 82.2 & 91.7 && 80.4 & 90.4 \\
{} & FBANK & 92.9 & 96.1 && 92.3 & 95.9 \\
\midrule
\multicolumn{7}{c}{\textbf{1 hour of labeled data}}\\
\midrule
\multirow{3}{*}{Discrete BERT} & vq-wav2vec & \textbf{8.5} & \textbf{16.4} && \textbf{9.0} & \textbf{17.6} \\
{} & MFCC & 14.1 & 32.1 && 14.3 & 32.5 \\
{} & FBANK & 14.2 & 30.6 && 14.6 & 31.7 \\
\midrule
\multirow{3}{*}{Continuous  BERT} & wav2vec & 22.1 & 42.0 && 22.4 & 44.0 \\
{} & MFCC & 53.8 & 75.0 && 52.8 & 74.5 \\
{} & FBANK & 56.7 & 76.2 && 55.0 & 76.1 \\
\midrule
\multicolumn{7}{c}{\textbf{10 hours of labeled data}}\\
\midrule
\multirow{3}{*}{Discrete BERT} & vq-wav2vec & \textbf{5.3} & \textbf{13.2} && \textbf{5.9} & \textbf{14.1} \\
{} & MFCC & 9.8 & 26.6 && 9.9 & 27.8 \\
{} & FBANK & 9.8 & 25.7 && 10.1 & 26.6 \\
\midrule
\multirow{3}{*}{Continuous BERT} & wav2vec & 13.6 & 31.7 && 14.1 & 34.3 \\
{} & MFCC & 27.5 & 54.2 && 27.4 & 55.7 \\
{} & FBANK & 25.0 & 50.2 && 24.9 & 51.7 \\
\midrule
\multicolumn{7}{c}{\textbf{100 hours of labeled data}}\\
\midrule
\multirow{3}{*}{Discrete BERT} & vq-wav2vec & \textbf{4.0} & \textbf{10.9} && \textbf{4.5} & \textbf{12.1} \\
{} & MFCC & 7.6 & 24.2 && 7.8 & 24.4 \\
{} & FBANK & 7.2 & 22.8 && 7.8 & 23.3 \\
\midrule
\multirow{3}{*}{Continuous BERT} & wav2vec & 11.3 & 26.4 && 11.8 & 28.3 \\
{} & MFCC & 11.6 & 34.0 && 12.4 & 35.5 \\
{} & FBANK & 10.1 & 30.9 && 11.0 & 31.8 \\
\bottomrule
\end{tabular}
\caption{%
Librispeech WER of BERT models taking discretized or continuous input features:
for discretized BERT quantized input units are obtained via vq-wav2vec or by k-means clustering of MFCC/FBANK features; for continuous BERT wav2vec features are learned following~\cite{schneider2019wav2vec}.
Pre-trained models are fine-tuned on various sizes of labeled data.
}
\label{tbl:libri-results}
\end{table}

We use the publicly available wav2letter++ \cite{w2l_pp} decoder integrated into the Fairseq framework with the official Librispeech 4-gram language model.
We run a sweep on weights for language model score, word score and silence token weights for each model, where parameters are chosen randomly and evaluated on the dev-other Librispeech set. We use the weights found by these sweeps to evaluate and report results for all other splits. The sweeps are run with beam size of 250, while the final decoding uses a beam size of 1500.


\begin{table}[t]
\centering
\setlength\tabcolsep{1.1pt}
\begin{tabular}{lccc}
\toprule
 & Labeled & \multicolumn{2}{c}{test}\\
 \cline{3-4}
& data & clean & other \\
\midrule
Wang et al. (2019) \cite{yongqiang_19} & 960h & 2.6 & 5.6 \\
\midrule
Irie et al. (2019) \cite{Irie_2019} (No LM) & 100h & 12.9 & 35.5 \\
Kahn et al. (2019) \cite{kahn2019selftraining} (Conv LM) & 100h & 5.9 & 24.1 \\
Panayotov et al. (2015) \cite{librispeech} & 100h &  6.6 & 22.5 \\ 
Lüscher et al. (2019) \cite{L_scher_2019} & 100h & 5.8 & 18.6 \\
\midrule
Kawakami et al. (2019) \cite{kawakami2019cpc8k} & 96h & 9.4 & 26.8 \\
\midrule
\multirow{4}{*}{vq-wav2vec + Discrete BERT (Ours)} & 10m & 16.3 & 25.2 \\
{} & 1h & 9.0 & 17.6 \\
{} & 10h & 5.9 & 14.1 \\
{} & 100h & 4.5 & 12.1 \\
\bottomrule
\end{tabular}
\caption{Comparison to previously published results in terms of WER on Librispeech. All results use 4-gram language models, except~\cite{Irie_2019, kahn2019selftraining}.}
\label{tbl:prev-results}
\end{table}

\subsection{Results}
\label{sec:results}

In our first experiment, we compare unit discovery followed by BERT training over the resulting discrete units (Discrete BERT) to directly learning representations from the audio inputs (Continuous BERT) in different simulated labeled data scenarios ranging from 100 hours to 10 minutes.
We compare quantization with vq-wav2vec to clustered MFCC and FBANK features.
The continuous BERT variant learns directly from the audio representations without explicit quantization and we experiment with inputting wav2vec, MFCC and FBANK features.

Table \ref{tbl:libri-results} compares WERs of different input features and pre-training methods on the standard Librispeech clean and other subsets. 
The first observation is that Discrete BERT outperforms Continuous BERT in all settings.
This shows that pre-training over meaningful discrete units outperforms directly learning representations from the continuous unlabeled data.
The initial unit discovery builds a vocabulary that makes the subsequent BERT pre-training more effective.

The best input features are obtained through self-supervised learning through vq-wav2vec \cite{baevski2019vqwav2vec} for Discrete BERT, or wav2vec \cite{schneider2019wav2vec} for Continuous BERT. 

For Discrete BERT, vq-wav2vec provides about 40\% of relative error reduction for both test sutsets compared to clustered spectral features across all training set sizes, with bigger gains on the noisy test-other subset. 

Pre-training brings clear benefits: When reducing the amount of labeled training data from 100h to 10h results in an increase of only 2 WER on test-other and 1.4 WER on test-clean for Discrete BERT with vq-wav2vec inputs. 
This shows that pre-training is effective and particularly so when little labeled data is available.
When reducing the amount of labeled data to only 10 minutes, Discrete BERT with va-wav2vec inputs can still achieve a WER of 16.3/25.2 on test-clean/other.

Table \ref{tbl:prev-results} shows a comparison of Discrete BERT to results from the literature. 
Fine-tuning Discrete BERT on only 10 hour of labeled data can nearly match the best known result on 100 hours of labeled Librispeech data~\cite{L_scher_2019} on test-clean, while achieving a 25\% relative WER reduction on test-other. Moreover, when using the same train-clean-100 subset for fine-tuning, Discrete BERT with vq-wav2vec inputs improves by 6.5 WER (35\% relative WER reduction) on test-other and 1.3 WER (22\% relative WER reduction) on test-clean over~\cite{L_scher_2019}.

The closest setup to ours is \cite{kawakami2019cpc8k} which learns representations using CPC~\cite{cpc} and then feed these into an ASR system trained on about 96h of labeled data, which is surpassed on test-clean by our Discrete-BERT model trained on vq-wav2vec representations fine-tuned on 10mins and 1 hour.




\begin{table}[t]
\centering
\setlength\tabcolsep{4.0pt}
\begin{tabular}{lccccc}
\toprule
\multirow{2}{*}{\shortstack{Input\\features}} & \multicolumn{2}{c}{dev} && \multicolumn{2}{c}{test}\\
\cline{2-3}\cline{5-6}
{} & clean & other && clean & other \\
\midrule
\multicolumn{6}{c}{Discrete input (no BERT pre-training)} \\
\midrule
vq-wav2vec & 99.7 & 99.3 && 99.3 & 99.4 \\
MFCC & 100.0 & 100.0 && 99.9 & 100.0 \\
FBANK & 98.8 & 99.6 && 98.9 & 99.0 \\
\midrule
\multicolumn{6}{c}{Continuous input (no BERT pre-training)} \\
\midrule
wav2vec & 27.5 & 47.2 && 28.4 & 48.5 \\
MFCC & 30.6 & 56.7 && 32.2 & 58.2 \\
FBANK & 20.2 & 47.6 && 21.3 & 49.2 \\
\bottomrule
\end{tabular}
\caption{%
WER on Librispeech with no pre-training for continuous and discrete input features on 100 hours of labeled data.
}
\label{tbl:nopre-results}
\end{table}

\begin{table}[t]
\centering
\setlength\tabcolsep{3.5pt}
\begin{tabular}{lrrrrr}
\toprule
\multirow{2}{*}{Pre-training} & \multicolumn{2}{c}{dev} && \multicolumn{2}{c}{test}\\
\cline{2-3}\cline{5-6}
{} & clean & other && clean & other \\
\midrule
\shortstack{wav2vec} & 51.1 & 71.4 && 52.6 & 71.9 \\
\shortstack{+ Continuous BERT} & 13.3 & 31.9 && 14.0 & 35.0 \\
\bottomrule
\end{tabular}
\caption{%
Comparison of single-step pre-training (wav2vec) to two-step pre-training (wav2vec + Continuous BERT). wav2vec is fine-tuned on 10 hours of labeled data and so is the Continuous BERT model with wav2vec inputs.
}
\label{tbl:w2v-ablation}
\end{table}

\subsection{Ablations}

To better understand the impact of BERT pre-training in our representation learning approach, we remove the BERT pre-training step and only perform unit discovery through vq-wav2vec, for discrete inputs, and fine-tuning, for both discrete and continuous inputs on the 10 hour labeled setup. 
The vocabulary for discrete inputs is still built on the unlabeled data.
Table~\ref{tbl:nopre-results} shows that training with discrete inputs fails. This is likely because the representations of the input discrete units are random and training on the labeled data is not sufficient. Continuous inputs do not suffer from this issue.


Next, we shed some light on how a two-step pre-training approach compares to a single-step approach. 
Specifically, we compare Continuous BERT with wav2vec input features (requiring separate learning of the wav2vec features) to just wav2vec features fine-tuned with a CTC loss on labeled data.
The results (Table~\ref{tbl:w2v-ablation}) show that Continuous BERT + wav2vec provides substantial gains. A second step of representation learning more than halved the WER, with more gains observed in the ``clean'' subset (cf. \ref{sec:results}).


\section{Discussion and Related Work}

The success of BERT~\cite{bert} and Word2Vec~\cite{mikolov2013word2vec} for NLP tasks motivated more research on self-supervised approaches for acoustic word embedding and unsupervised acoustic feature representation \cite{sbengio_14, Livescu_2013, audio_w2v, Wanjia_16, s2v, picheny_19, schneider2019wav2vec, cpc, glass_19_cpc, baevski2019vqwav2vec}, either by predicting masked discrete or continuous input, or by contrastive prediction of neighboring or similarly sounding segments using distant supervision or proximity in the audio signal as an indication of similarity. In~\cite{goldwater_15} a dynamic time warping alignment is used to discover similar segment pairs. 

Our work is inspired by research efforts reducing the dependence on labeled data for building ASR systems through unsupervised unit discovery and acoustic representation leaning \cite{pg_08, aren_10, glass_12, JHU_2012, JSALT_2017}, and through multi- and cross-lingual transfer learning in low-resource conditions \cite{Haihua_2016, jia_2015, Georg_2013, arnab_2013, Huang_2013, Ngoc_2014}, and semi-supervised learning \cite{Vesely_13, Sheng_2017, hari_2019, tara_sslearning}.

\section{Conclusion and Future work}

We presented a systematic comparison of self-supervised pre-training approaches for speech recognition.
The most effective method is to first learn a discrete vocabulary of the data with vq-wav2vec followed by standard BERT training over these discrete units. 
This performs much better than directly learning from the continuous audio data.
Different to previous work which relied on task-specific ASR models, we directly fine-tune the resulting BERT model on transcribed speech data to act as speech recognition models.
This approach can achieve better accuracy on test-other than the best known result with 100 hours of labeled data while relying on two orders magnitude less labeled data.
When the model is fine-tuned on only 10 minutes of data, it can still achieve a WER 25.2 on test-other and WER 16.3 on test-clean.


\footnotesize
\bibliography{refs}
\bibliographystyle{IEEEbib}

\end{document}